\documentclass[conference]{IEEEtran}
\IEEEoverridecommandlockouts
\usepackage{cite}
\usepackage{amsmath,amssymb,amsfonts}
\usepackage{algorithmic}
\usepackage{graphicx}
\usepackage{textcomp}
\usepackage{xcolor}
\def\BibTeX{{\rm B\kern-.05em{\sc i\kern-.025em b}\kern-.08em
    T\kern-.1667em\lower.7ex\hbox{E}\kern-.125emX}}
\begin{document}

\title{Optimising Design Verification Using Machine Learning: An Open Source Solution\\
{}
\thanks{}
}

\author{\IEEEauthorblockN{1\textsuperscript{st} Samhita Varambally B.}
\IEEEauthorblockA{\textit{National Institute of Technology Karnataka, Surathkal} \\
samhitav@gmail.com}
\and
\IEEEauthorblockN{2\textsuperscript{nd} Naman Sehgal}
\IEEEauthorblockA{\textit{Birla Institute of Technology and Science, Pilani} \\
namansehgal95@gmail.com}

}

\maketitle

\section{Introduction}
Design Verification is currently the bottleneck in the ASIC Design Flow. The verification flow involves several simulations which are tested in regressions with the goal of exercising all the critical scenarios in the design. These scenarios are defined by coverage metrics which need to hit 100\% for coverage closure. With the growing complexity of SoCs, the industry has reached a stage where double the number of verification engineers are required for every design engineer, and the verification cycle significantly impacts time to tape-out of the chip. The cost of verification is fast approaching half the total cost of the chip design\cite{olofsson2017intelligent}. Hence the question arises, what can we do to make this process more efficient, faster and less expensive? In this work, we present a method which employs machine learning to increase efficiency in catching bugs and speeds up the complete verification cycle by accelerating coverage convergence. The method uses supervised learning with Artificial Neural Networks to achieve the same. A software centred approach to verification is chosen for this, on an open source platform called Cocotb.
%The slow growth and adoption of Hardware Verification Languages (HVL) such as SystemVerilog and Specman E make it difficult to build complex applications quickly. A software centred approach to functional verification seems to be more intuitive. Thus we adopted an open source based framework, Cocotb, in our methodology. This uses Python for testbench creation which is far more robust and quicker to deploy[].

\section{Background and Related Work}
Traditional directed tests have been replaced in the industry with Coverage driven Constrained Random Stimuli, as directed tests on the increasingly large and complex SoCs being designed are unrealistically complicated and time consuming\cite{mehta2018constrained}. This method involves a functional coverage model defined by the engineer at the start of the verification cycle from the verification plan to cover all features and corner cases in the design. The industry standard for functional design verification is the use of Hardware Verification Languages(HVL), which are derived from Hardware Description Languages(HDL) such as Verilog or VHDL. These HVLs are complicated and rigid, for example, SystemVerilog comprises of 221 keywords and the Universal Verification Methodology which is built on System Verilog is even more complex with over 300 classes\cite{rosser2018cocotb}\cite{uvm}.\\%They are better suited for hardware design and not for verification.\\
A testbench comprising of random stimulus generation, signal monitoring and checking is built using a HVL. Constrained Random Verification generates a random stimulus given a unique input seed for each simulation. The simulations are repeated in regressions with random seeds until full coverage closure is achieved. This is not efficient as the random stimuli would take a large number of iterations to hit obscure corner cases in the design. Some cases may never be hit even after several regressions, which would then require effort in manipulating the input constraints to change stimuli, analysing and prioritising the unverified features of the design by the verification engineer. Finding the right combination of constraints to simulate the most stressful tests on the design is a challenge which is time consuming. Moreover, even if all coverage parameters are hit, design critical corner cases may be exercised only once while simple and straightforward scenarios may be exercised most of the time due to the random nature of the stimulus in regressions. For example, if a  functionality of the design is accessed only when a large number of signals all occur at once, that functionality would be scarcely tested when inputs are randomised and may require manual updation of input stimuli to simulate such a scenario. This could lead to bugs being undetected throughout the verification cycle as it is not possible to manually generate a large number of difficult to hit test cases. %Furthermore, when a bug is detected in the design, it would be necessary to find and test more such similar tests and exercise that feature of the design thoroughly. This process heavily relies on the intellect and efficiency of the verification engineer to be able to identify similar scenarios in the design and simulate multiple tests exercising the buggy code. A machine learning approach would enable a more thorough testing of corner cases.

\section{Approach and Uniqueness}
The approach used in this research is unique due to two main differences over the traditional methods.
\subsection{Open Source Software Oriented Approach}
Currently, there are no free or open source compilers available for HVLs. To verify a digital integrated circuit design, one has to invest in licenses which are expensive and add a significant cost when multiple regressions are being simulated. Moreover, traditional methods of creating testbenches using SystemVerilog and UVM are time consuming and have a steep learning curve for new engineers. Using a higher level, general purpose programming language is more practical for verification. Cocotb is an open source testbench environment for verifying Verilog and VHDL design. It is built with the aim of lowering the overhead of creating a test \cite{cieplucha2017new}. Cocotb uses Python to program testbenches which offer several advantages for our use case.
Firstly, it is simple with only 23 keywords, intuitive and robust. Secondly, the plethora of online resources and libraries of existing code make it easy for a novice engineer to code and debug when compared to System Veilog or Specman E. Python is ideal for machine learning applications due to the extensive work done on the same and existing open source libraries which can be plugged in to build machine learning models quickly which are unavailable in HVLs.
%\begin{itemize}
 %   \item Simple(only 23 keywords), intuitive and robust, possible to model complex verification scenarios quickly
  %  \item Has a huge library of existing code which can be reused and active online forums to solve issues
   % \item Python is popular, most engineers already know it, or can learn it quickly when compared to System Verilog or Specman E
    %\item Python is ideal for machine learning applications due to a plethora of existing libraries and online resources
%\end{itemize}

In this research, Cocotb is proposed to model the testbench in Python and hardware description languages like Verilog is restricted to the design. Cocotb requires a simulator to simulate only the HDL design for which an open source compiler, Icarus Verilog can be used. This completely eliminates the need for licenses.\cite{cocotb}

\subsection{Machine Learning for Coverage Closure and Efficient Bug Tracking}
In Constrained Random Verification, one often has to manually update constraints in order to target coverage holes and hit corner cases. Furthermore, when a bug is detected in the design it is necessary to find and test more such similar scenarios and exercise that feature of the design thoroughly. This process heavily relies on the verification engineer's understanding of the design and his efficiency in creating multiple test cases for rigorous verification. Apart from the issues raised in Secion 2, these human factors have the potential to adversely impact the efficiency of stimulus generation and the quality of coverage. Abstracting the need to comprehend the design for verification requires automatic updating of constraints. This can be done with a machine learning based approach which will enable a more exhaustive testing of corner cases. The key to do this is to use functional coverage and test result as a feedback to the system. Coverage metrics can convey device configurations used, cases covered and more importantly, scenarios not yet covered in verification. Inputs are then automatically modelled to hit non covered scenarios. 

The process for machine learning based input stimulus generation is as follows:
\subsubsection{Coverage Based Approach}
Supervised learning is when a machine learning model is given a large set of inputs and outputs (training dataset) considered ground truth. The model maps the outputs as a function of the inputs. Once the model is trained, it can be used to predict the outputs for any combination of inputs to the model\cite{zhu2009introduction}. The intriguing feature of the methodology used in this approach is that the training dataset is generated by the environment itself. This mitigates the need for external datasets or design knowledge to train the model. This is done as follows
\begin{itemize}
\item A number of tests are simulated with random stimulus. For each simulation, the inputs to the the device under test (DUT), outputs from the DUT, coverage information and test pass or fail status is captured.
\item The captured data is used as the labelled training dataset to train an Artificial Neural Network(ANN) which models the verification inputs as a function of the coverage information\cite{hassoun1995fundamentals}. 
\item These simulations form the training dataset for the Neural Network, and eventually the model is able to predict the verification inputs which can cover a given functional coverage metric.
\item The coverage model in the testbench can then be changed to include only the corner cases or rare scenarios and exclude the simple, basic parameters which are most often exercised in random regressions.
\item The ANN model is then used to predict inputs which lead to these obscure corner cases in the test phase, thus increasing the number of tests hitting critical design while regressions are fired. 
\end{itemize}
Once the ANN model is trained, regressions, which are a series of simulations are fired with ML based stimulus. This stimulus is got by feeding in the coverage data of the previous simulation to the ANN model. The model predicts inputs which would cover the holes in the coverage database.
\subsubsection{Test Failure Based Inputs}
An extension to the above method is to directly target bugs in the design by analysing a simulation failure similar to uncovered coverage parameters in the above section. This will enable the model to learn how to generate verification inputs which may lead to such design failures. Over time, such a verification environment will move in a direction to repeatedly produce the scenarios which may lead to the discovery of more design bugs. Furthermore, if a failure is found, this method will lead to thorough verification of design functionality which generated such a failure.
\subsection{Evaluation}
\begin{figure}[htbp]
\centerline{\includegraphics[scale=0.38]{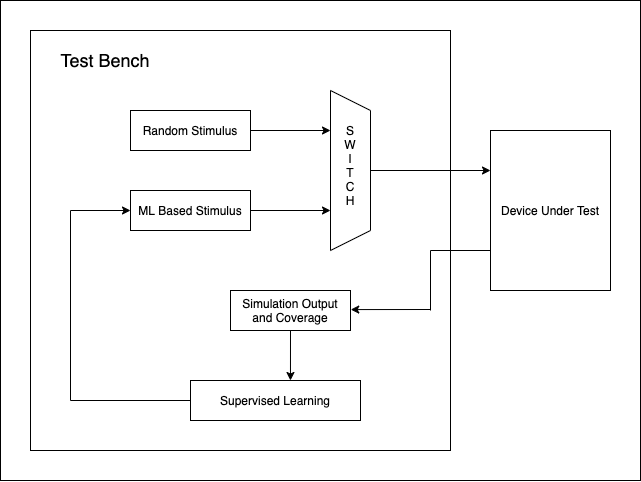}}
\caption{Block Diagram of the Verification Environment}
\label{packet}
\end{figure}
The above figure shows the environment used to evaluate the model. The Supervised Learning is accomplished using an ANN model. Random stimulus is used in the training phase and a multiplexer is used to switch to the ML based inputs received through feedback from the model during the verification test phase. The DUT used to test this methodology is a comparator designed in Verilog with varying widths to check the number of iterations for coverage convergence for different widths. Regressions are run continuously till coverage convergence is achieved. \\The results using traditional Constrained Random Verification and the Coverage Based Machine Learning Approach is tabulated below.

\section{Results}
\begin{table}[htbp]
\caption{Comparison of Coverage Convergence with Traditional Method vs ML Based Approach}
\begin{center}
\begin{tabular}{|c|c|c|}
\hline
\textbf{}&\multicolumn{2}{|c|}{\textbf{Number of Iterations}} \\
\hline
%\cline{2-5} 
\textbf{Width}&\textbf{Constrained Random Stimuli}&\textbf{ANN Based Stimuli}\\
\hline
1&3&3 \\
\hline
2&60&16 \\
\hline
3&261&24 \\
\hline
4&segmentation fault&29 \\
\hline
5&segmentation fault&96 \\
\hline

\end{tabular}
\label{results}
\end{center}
\end{table}

As observed in the above table, ANN based approach significantly reduces the number of iterations for coverage convergence. The traditional method gives a segmentation fault for comparator widths of 4 and 5, which implies that coverage closure could not be achieved for the design without manual intervention for stimulus updation\cite{ambalakkat2018optimization}.
\section{Contributions}
Machine learning inputs are seen to perform better than random inputs for verification of complex ICs. This methodology is proved to greatly reduce the time for finding hard to hit scenarios and generating obscure testcases for coverage closure. 
In a verification cycle, regressions are simulated multiple times for every iteration of the RTL design phase. Regressions are used to verify Gate Level Simulations(GLS) as well, which are known to be far more time consuming than RTL simulations. Faster coverage closure will thereby exponentially reduce the total time for verification over a verification cycle from RTL to GLS, thus leading to quicker tapeout of the chip.\\
Furthermore, each regression is be more thorough due to exercising of corner cases more often and intelligently simulating probable bugs in the design. The efficiency of bug tracking is also improved due to intelligent generation of similar scenarios leading to exhaustive testing of the buggy design. This ensures rigorous verification of critical design scenarios which is not accomplished through Constrained Random Verification. This method also reduces the resource requirement of verification engineers by reducing the need for manual intervention. 
\section{Drawbacks}
This methodology is seen to work best for small designs and ICs. It is not yet viable to deploy such a solution on a large and complex SoC design. The test phase is extremely long in such a case and one test case per feature may be sufficient to call verification closure. Machine learning based inputs help in the case of rigorous testing involving multiple tests for each feature of the design. Another drawback is that even with this approach manual intervention is necessary, or rather more efficient to generate corner cases for dynamic designs which have regular feature updates or multiple feature additions done in a design cycle. As design and verification often proceed in parallel, manual intervention becomes unavoidable in such a case. 

\bibliographystyle{./bibliography/IEEEtran}
\bibliography{./bibliography/IEEEabrv,./bibliography/IEEEexample}

\end{document}